\documentclass{article}
\usepackage{ijcai22}

\usepackage{amssymb}
\usepackage{rotating}
\usepackage{array}
\usepackage{subfig}
\usepackage{times}
\usepackage{soul}
\usepackage{url}
\usepackage[hidelinks]{hyperref}
\usepackage[utf8]{inputenc}
\usepackage{caption}
\usepackage{graphicx}
\usepackage{amsthm}
\usepackage{booktabs}
\usepackage{algorithm}
\usepackage{multirow}
\usepackage{amsmath}
\usepackage{algpseudocode}
\usepackage{algorithmicx}
\usepackage{color}
\usepackage{xcolor}
\usepackage{textcomp}

\newcommand{\tabincell}[2]{\begin{tabular}{@{}#1@{}}#2\end{tabular}}
\title{Discriminative Feature Learning Framework with Gradient Preference for Anomaly Detection}

\author{
MuHao Xu$^{1,2}$
\and
Xueying Zhou$^{1,2}$\and
Xizhan Gao$^{1,2}$\And
 WeiKai He$^{3}$\And
Sijie Niu$^{1,2,}$
\footnote{Corresponding Author, Muhao Xu and Xueying Zhou contributed to the work equally and should be considered as co-first authors.}
\affiliations
$^1$School of Information Science and Engineering, University of Jinan, Jinan 250022, China\\
    $^2$  Shandong Provincial Key Laboratory of Network-based Intelligent Computing, Jinan 250022, China\\
$^3$ ShanDong Jiaotong University. Jinan 250022, China\\
\emails
sjniu@hotmail.com,
}

\begin{document}

\maketitle

\begin{abstract}
Unsupervised representation learning has been extensively employed in anomaly detection, achieving impressive performance. Extracting valuable
feature vectors that can remarkably improve the
performance of anomaly detection are essential in
unsupervised representation learning. To this end,
we propose a novel discriminative feature learning
framework with gradient preference for anomaly
detection. Specifically, we firstly design a gradient
preference based selector to store powerful feature
points in space and then construct a feature repository, which alleviate the interference of redundant
feature vectors and improve inference efficiency.
To overcome the looseness of feature vectors, secondly, we present a discriminative feature learning with center constrain to map the feature repository to a compact subspace, so that the anomalous
samples are more distinguishable from the normal
ones. Moreover, our method can be easily extended
to anomaly localization. Extensive experiments
on popular industrial and medical anomaly detection datasets demonstrate our proposed framework
can achieve competitive results in both anomaly
detection and localization. More important, our
method outperforms the state-of-the-art in few shot
anomaly detection.

\end{abstract}

\section{Introduction}
%Anomaly detection of products to ensure product quality and optimise the user experience is a very important part of the process of industrial intelligence.
%Visual inspection is a well-known, flexible and non-contact technique. has been widely applied to anomaly detection.

Anomaly detection is to assess whether an observed sample is normal or anomalous.
It has been received ever-increasing attention in the field of computer vision with wide applications in various scenarios, such as automatic driving field \cite{li2020multi}, disease diagnosis \cite{li2018thoracic}, and industrial defects detection \cite{bergmann2019mvtec}, et al.
Generally, challenges arising in anomaly detection are as follows:
i) As the types of anomalies are complicated and various, it is difficult to collect all of them;
ii) The number of anomalous samples in the vast majority of datasets is very rare.
As a consequence, unsupervised methods are becoming mainstream anomaly detection techniques.

Unsupervised anomaly detection methods are desirable to train models solely on anomaly-free images.
Existing works predominantly focus on reconstruction-based methods, such as  Autoencoder \cite{akcay2018ganomaly,bergmann2019mvtec,nguyen2019anomaly} or Generative Adversarial Networks (GAN) \cite{schlegl2019f}.
The principle of these methods is to detect anomalies with per-pixel reconstruction error or by evaluating the density obtained from the model's probability distribution.
Although features learned by the reconstruction-based methods are beneficial to pixel reconstruction, it is difficult for these methods to capture features conducive to anomaly detection.
 \begin{figure}[tbp]
  \centering
  \includegraphics[width=0.48\textwidth]{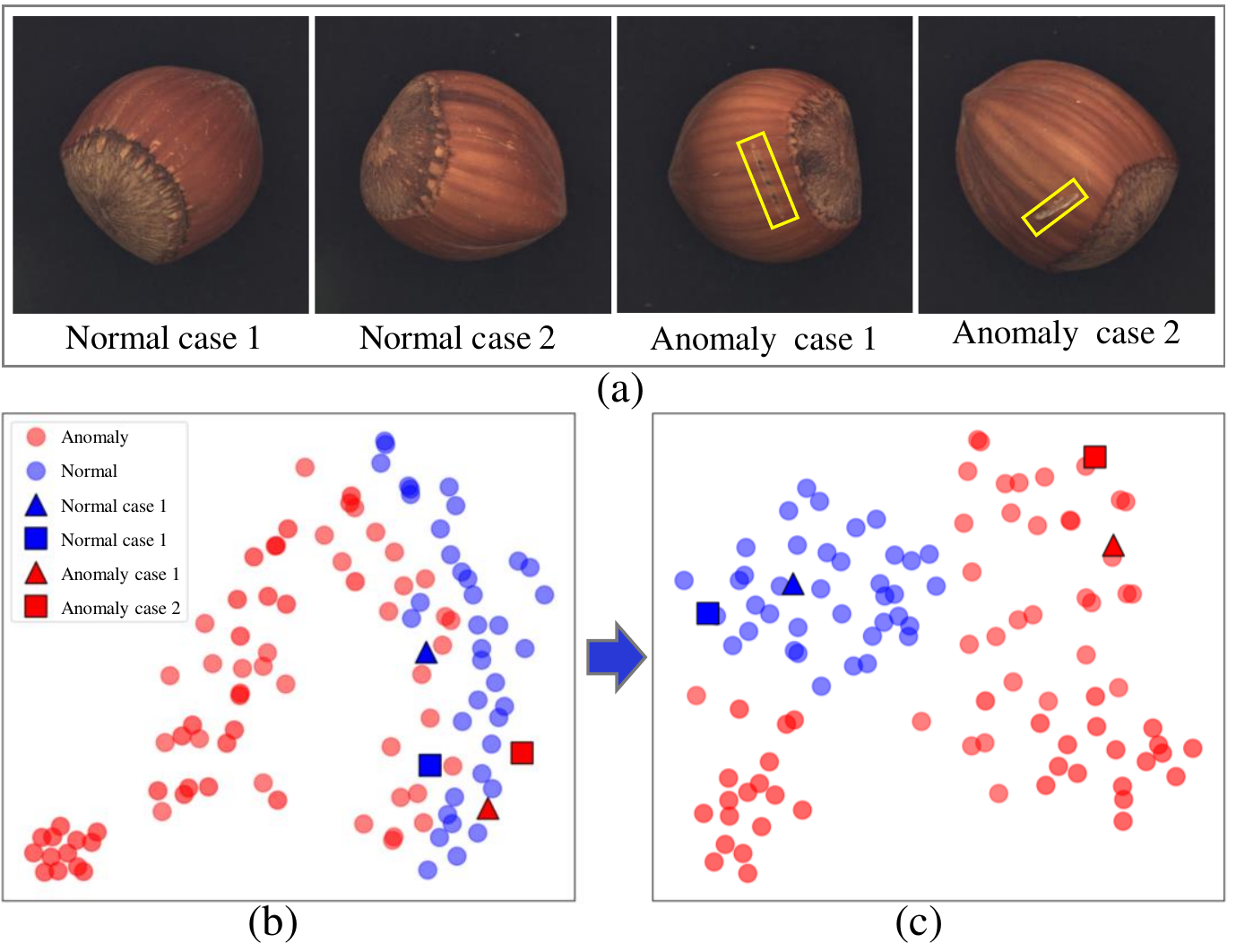} %1.png是图片文件的相对路径
  \caption{Visualization of feature distribution with and without mapping via t-SNE. (a) examples of normal and abnormal cases, (b) distribution of features extracted from the pre-trained network on test data, (c) features distribution of our proposed method on test data.} %caption是图片的标题
   \label{imgf1}%此处的label相当于一个图片的专属标志，目的是方便上下文的引用
\end{figure}
Recently, representation-based methods extract feature vectors or patches to distinguish anomalies,
achieving promising results. To enable better learning image semantics in a limited number of samples, pre-trained networks were introduced into anomaly detection methods \cite{reiss2021panda}, and their performance has been turned out to be significant improvement in anomaly detection.
PaDiM \cite{defard2021padim} used multivariate Gaussian distributions to describe the features extracted from the pre-trained networks, and then employed Mahalanobis distance metric as an anomaly score.
CutPaste \cite{li2021cutpaste} learned self-supervised deep representations from their proposed augmentated dataset, and build a generative classifier for anomaly detection.
FYD \cite{zheng2021focus} proposed a coarse-to-fine non-contrastive learning method to make the distribution of normal images become more compact and semantically meaningful, so that the outliers are more feasible to be captured.

Although the previous works have achieved considerable success, some challenging problems still need to be addressed:
i) most of the defects, as shown in Figure \ref{imgf1} (a), can cause high gradient values in corresponding defect regions. Characterization of these gradient are conductive to anomalous regions detection.
To our best knowledge, existing works only employ gradient information to get anomaly score \cite{liu2020towards}, which possible limit the generalization potential of the inference process;
ii) representation-based methods usually consist of feature extraction with pre-trained networks and anomaly detection using the extracted features.
Existing methods mainly focus on refining the pre-trained networks for extracting features adapted to the target distribution.
However, these pre-tasks are far away from final anomaly detection, leading to limited improvement.

To mitigate the aforementioned issues,
we propose a discriminative feature learning framework with gradient preference for anomaly detection.
In our observation, anomaly regions can be easy to cause large changes in image intensity. Thus, we employ gradient information to preserve valuable feature vectors from those obtained by pre-trained network, which significant improves the quality of feature representation of normal samples. Then, different from previous works that detect anomalies using the extracted feature vectors directly, we construct a mapping function to constrain them to a compact subspace, so that the obtained feature vectors are more discriminative, as depicted in Figure \ref{imgf1} (c). Moreover, our proposed framework can be easily extended to anomaly localization. Our main contributions are summarized as follows:
\begin{itemize}
\item We propose a gradient-preference based feature selection strategy to store powerful feature points in space and then construct a feature repository, which alleviates the interference of redundant feature vectors and improve inference efficiency.
\item We construct a discriminative feature learning network to make the feature repository of normal samples so compact that the normal and abnormal samples are more separable.  Furthermore, an early stop strategy is given to avoid the ``mode collapse'' problem.
\item Extensive experiments are conducted on four popular industrial and medical anomaly detection datasets. The results demonstrate that our proposed method can achieve competitive in both anomaly detection and localization.
\end{itemize}

\section{Related Work}
In this section, we review the related works on reconstruction-based and representation-based anomaly detection methods.

\textbf{Reconstruction-based} anomaly detection methods rely on encoder-decoder scheme, and use the reconstruction error as a metric to detect anomalies.
Akçay et al. \cite{{akccay2019skip}} proposed an encoder-decoder convolutional neural network with skip connections, and detect anomalies by combining feature and image reconstruction errors.
Schlegl et al. \cite{schlegl2019f}  trained an additional encoder after having trained a Generative Adversarial Network (GAN) so that the abnormal images can be detected  due to large reconstruction errors.
Liu et al. \cite{liu2020towards} localized abnormal regions in images by means of gradient-based attention maps generated from Variational Auto-Encoder (VAE).

\textbf{Representation-based} anomaly detection methods extract feature vectors or patches from images, and detect anomalies by the representation distance between the test images and normal cases.
MKD \cite{salehi2021multiresolution} distilled the knowledge of normal samples from the pre-trained teacher network into a more compact student network, which makes the network concentrating solely on the abnormal samples with obvious differences between the teacher network and the student network.
Ammar et al. \cite{mansoor2021anomaly} modeled local/point pattern features as a random finite set (RFS), and used their proposed RFS energy as anomaly score to detect anomalies.
Furthermore, VT-ADL  \cite{mishra2021vt} constructed an adapted vision transformer network, which is later processed by a Gaussian mixture density network to localize the anomalous regions.
CS-Flow \cite{rudolph2021fully} proposed a novel fully convolutional cross-scale normalizing flow (CS-Flow) that jointly processes multiple feature maps of different scales and detects anomalies by jointly estimating likelihoods on multiscale feature maps.

\nocite{golan2018:deep}
\nocite{akcay2018ganomaly}
\nocite{zhai2016deep}
%\nocite{nazare2018pre}
\nocite{rudolph2021same}
\nocite{bergmann2020uninformed}
\nocite{reiss2021panda}
\nocite{salehi2021multiresolution}
\nocite{defard2021padim}
\nocite{li2021cutpaste}
\nocite{mishra2021vt}
\nocite{yi2020patch}
\nocite{zavrtanik2021draem}
\nocite{mansoor2021anomaly}
\nocite{bergmann2019mvtec}
\nocite{huang2020surface}
\nocite{nguyen2019anomaly}
\nocite{schlegl2019f}
\nocite{li2020multi}
%\nocite{gudovskiy2021cflow}
\nocite{rudolph2021fully}
\nocite{li2018thoracic}
\nocite{liu2020towards}
\nocite{zheng2021focus}
\nocite{akccay2019skip}
\nocite{bergmann2018improving}
\nocite{sheynin2021hierarchical}
\section{Method}
\begin{figure*}
  \centering
  \includegraphics[width=\textwidth]{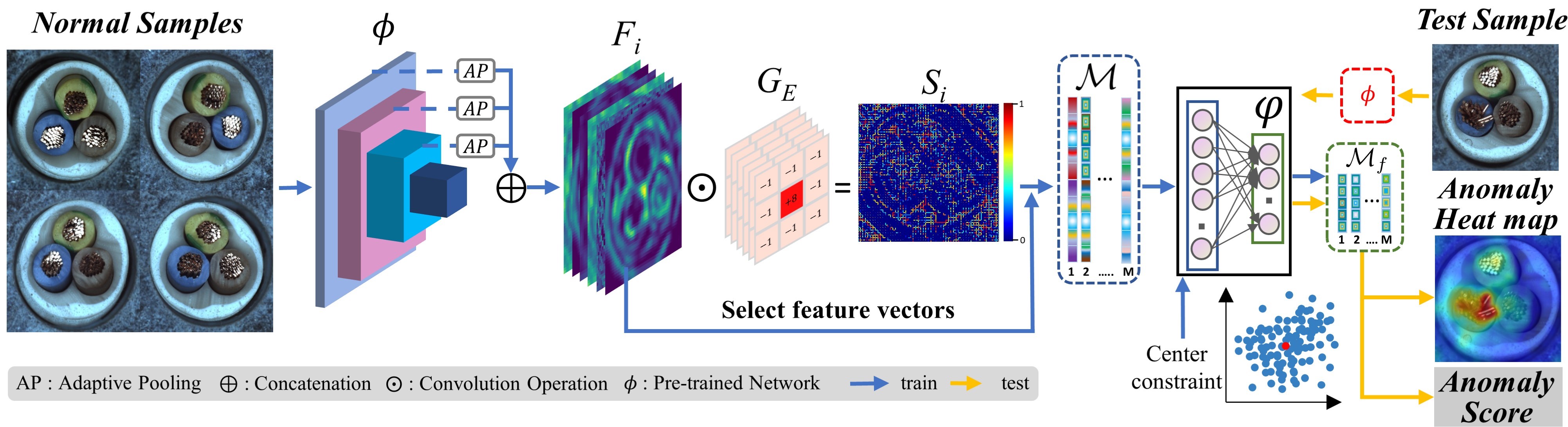} %1.png是图片文件的相对路径
  \caption{Overview of our framework for anomaly detection.} %caption是图片的标题
  \label{img1} %此处的label相当于一个图片的专属标志，目的是方便上下文的引用
\end{figure*}

To make the normal and anomalous samples more distinguishable, it is necessary to consider a discriminative learning network for bringing the features obtained by pre-trained network close to the better anomaly detection results. In this paper, we propose a method for selecting powerful feature vectors from the outputs of the mapping function by gradient preference and discriminative feature learning.
The architecture of our proposed method is illustrated in Figure \ref {img1}.

\subsection{Feature Extraction with Pre-trained Network}

The pre-trained networks are able to capture multi-level features in a supervised method, which can be transfered to anomaly detection and achieve competitive performances.
In our work, we first obtain a pre-trained network $\phi$, such as ResNet and EfficientNet, trained on an additional dataset (e.g. ImageNet).
Then, our training images $x_{i}\in\mathbb{R}^{ c\times h \times w} $ (channel $c$, height $h$ and width $w$) are fed into the pre-trained network $\phi$, and multi-level features  $f_{i,l}$ are defined as:
\begin{align}
f_{i,l} = \phi (x_{i})
\end{align}
where $ l $ denotes different levels, and $x_{i}$ denotes $i$-th training sample.

In general, the low-level features represent the edge and texture of images, while the high-level features represent semantic information of images.
Hence, we combine features of different semantic levels to get the final representative feature $F_{i} \in \mathbb{R}^{ C\times H \times W} $ of each image $x_{i}$, this process can be written as:
\begin{align}
F_{i} = AP (f_{i,1}) \oplus AP (f_{i,2}) \oplus … \oplus AP (f_{i,l})
\end{align}
where $\oplus$ represents concatenation in channel dimension, and $AP(\cdot)$ represents adaptive pooling that can resize $f_{i,l}$ into the resolution of maximum one.
Finally, we obtain the feature vectors of pixel location $(j, k)$ in an image $x_{i}$, defined as $F_{i}(j,k) \in \mathbb{R}^{C}$ , where $j \in [0,W)$ and $k \in [0,H)$ .

\subsection{Gradient Preference based Feature Repository}

The feature vectors obtained by previous step can be directly employed to anomaly detection.
However, the accuracy is limited as the redundant features are far away from the center of feature space.
To overcome this problem, many feature selection strategies are proposed.
Unlike previous works using random or distance-based selection strategy, we take gradient values which are observed usually high in abnormal regions of an image into consideration to select feature vectors.
Thus, we propose a novel feature selection strategy based on gradient-preference to select feature vectors that are beneficial to anomaly detection.

Concretely, inspired by image sharpening, we calculate gradient on the feature maps to generate the score map $S_{i}\in\mathbb{R}^{ H \times W}$, defined as:
\begin{align}
S_{i}=\Xi (F_{i}\odot G_{E})
\end{align}
where $S_{i}$ indicates the gradient values of feature $F_{i}$, $\odot$ is the convolution operation, $\Xi (\cdot)$ is the normalization function, and  $G_{E}\in \mathbb{R}^{C \times 3\times 3}$ is the convolution kernel obtained by extending laplacian operator $G$ in channel dimension, and the operator $G$ is as follow:
\begin{align}
    G=\begin{bmatrix} -1 & -1 & -1 \\ -1 & 8 & -1 \\ -1 & -1 & -1 \end{bmatrix}
\end{align}

Most directly selection strategies are to sort score map from large to small and select feature vectors with high values, whereas in fact that feature vectors without high gradient values are also significant for anomaly detection.
Different from it, we select feature vectors by probability which can select feature vectors with high gradient values as well as those without.
Firstly, we generate a random tensor $R_{i} \in \mathbb{T}^{H \times W}$, where $\mathbb{T} \sim $ U$(0,1)$.
%Then, we binarize $S_{i}$ to $B_{i}$ with $R_{i}$ as the threshold, and this process can be written as:
Then, the mask $B_{i}$ based on gradient-preference is defined as the following expression:

If $B_{i}^{j,k}=1$, we store the corresponding powerful feature points in space.
Finally, according to the stored powerful feature points, we get the feature repository $\mathcal{M} \in \mathbb{R}^{M \times C}$ by:
\begin{align}
\mathcal{M}=\bigcup_{(j,k) \in \{(j,k)|B_{i}^{j,k}=1\}}{F_{i}(j,k)}
\end{align}

\subsection{Discriminative Feature Learning with Center Constraint}

Although the above steps make the feature space of normal samples more representative, there are still some problems to be solved. For one thing, the feature vectors we selected are redundant, which leads to inaccurate feature representation. For another thing, the feature vectors are so loose that the normal and abnormal samples are indistinguishable.

To alleviate the above problems, we utilize a simple Multilayer Perception (MLP) network $\varphi$ with central constraint loss to map the feature vectors into a compact subspace with $C_{f}$ dimensions, so that the features are more discriminative. Our proposed central constraint loss is as follow:
\begin{align}
L_{center}=\sum ^M_{1}\left|\left| \dfrac{1}M\sum ^M_{1}\varphi\left( v\right)-\varphi\left( v\right) \right|\right|_{1}; v \in \mathcal{M}
\end{align}

Under the condition of full training, the network $\varphi$ suffers "mode collapse" that all feature vectors are mapped to the same output and become inseparable.
To avoid that, we design an early stopping strategy.
Specifically, we set a hyperparameter $\eta$, and stop the optimization of the network $\varphi$ when the given condition $\mathcal{C}$ is met. The form of the condition $\mathcal{C}$ is as follow:
\begin{align}
L_{center}\leq\eta*L_{start}
\end{align}
where $L_{start}$ represents the loss values after training one epoch.
Finally, we obtain the final repository $\mathcal{M}_{f} \in \mathbb{R}^{M \times C_{f}}$ by:
\begin{align}
\mathcal{M}_{f}=\bigcup_{v \in \mathcal{M}}{\varphi(v)}
\end{align}
\subsection{Anomaly Detection in Inference Stage}
In inference phase,  we can detect and localize anomalies through the following procedures.
%In inference stage, our method achieves considerable performance in both anomaly detection and localization.

\textbf{Anomaly detection} refers to assessing whether a test image is abnormal or not.
To this end, we feed a test image into the feature extraction module, which yields $\small{W \times H}$ feature vectors $F_{test}(j,k)$.
Then, all feature vectors are fed into the trained MLP network $\varphi$, and the feature repository of the test image $\mathcal{M}_{test} \in \mathbb{R}^{(W \times H) \times C_{f}}$ without selection is obtained.

A high anomaly score indicates that the image is likely to be abnormal.
Since all the feature vectors stored in the repository $M_{f}$ are from normal images, the features of normal images are very close to those of repository.
In contrast, the features of abnormal images have obvious difference from those of normal images in the repository $M_{f}$, leading to high anomaly scores.

\textbf{Anomaly localization} refers to judging whether a pixel is in abnormal region or not.
Similarly, we calculate anomaly scores $\mathcal{A}_{pixel}$ for pixels to indicate the possibility that the pixels are in abnormal regions.
The anomaly scores $\mathcal{A}_{pixel}$ can be represented as:
\begin{align}
    \mathcal{A}_{pixel}&= \left\| v_{test}-v^{\star}_{f}\right\| _{2}
\end{align}

Then, we get an attention map $A \in \mathbb{R}^{h \times w}$, which concentrates on abnormal regions in an image.
Finally, we resize this attention map into the resolution of the original image by bilinear interpolation, and smooth it with a Gaussian kernel $ \sigma = 4$.
The pseudo-code for our proposed method is shown in Algorithm 1.

 \begin{figure*}
  \centering
  \includegraphics[width=1\textwidth,height=8.3cm]{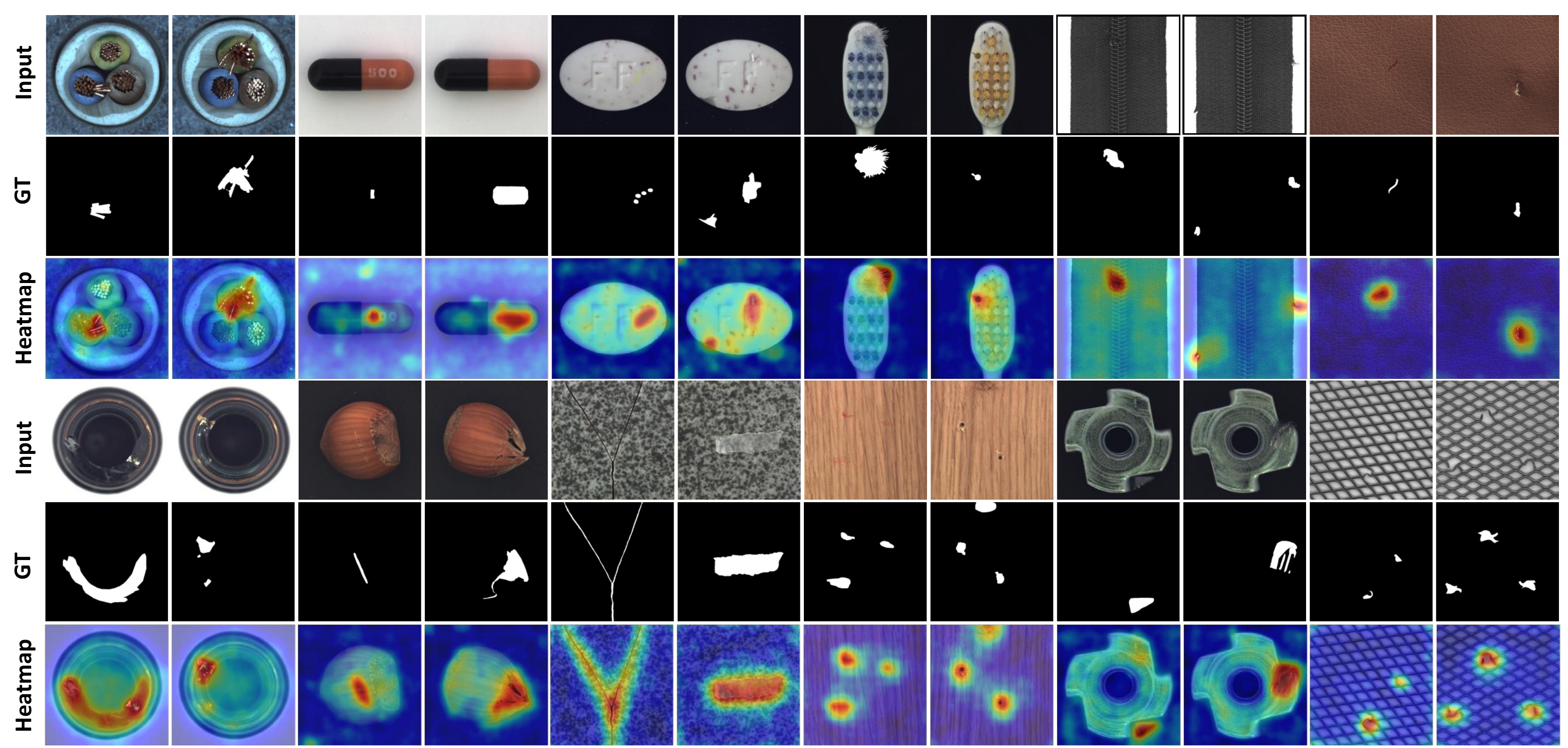} %1.png是图片文件的相对路径
  \caption{Visualization of anomaly localization results on different categories of  MVTec AD dataset, which are cable, capsule, pill, toothbrush, zipper, leather, bottle, hazelnut, tile, wood, metal nut and grid in turn. (From top to bottom and left to right).} %caption是图片的标题
  \label{img3} %此处的label相当于一个图片的专属标志，目的是方便上下文的引用
\end{figure*}

\begin{table*}[t]
\renewcommand\arraystretch{1.2}
  \centering
   \resizebox {\textwidth}{!}
   {
       \begin{tabular}{lrrrrr|rrrrrrrrrrr}
\hline
  \multicolumn{1}{l}{{Category}} & \multicolumn{5}{c|}{Textures}         & \multicolumn{11}{c}{Objects} \\
    \multicolumn{1}{l}{} & Carpet & Grid  & Leather & Tile  & Wood  & Bottle & Cable & Capsule & Hazelnut & Metal Nut & Pill  & Screw & Toothbrush & Transistor & Zipper & \textbf{Avg} \\

   \hline
    GeoTrans  \cite{golan2018:deep} & 43.7/\ \ -- \ \ \   & 61.9/\ \ -- \ \ \   & 84.1/\ \ -- \ \ \   & 41.7/\ \ -- \ \ \   & 61.1/\ \ -- \ \ \   & 74.4/\ \ -- \ \ \   & 78.3/\ \ -- \ \ \   & 67.0/\ \ -- \ \ \   & 35.9/\ \ -- \ \ \   & 81.3/\ \ -- \ \ \   & 63.0/\ \ -- \ \ \   & 50.0/\ \ -- \ \ \   & 97.2/\ \ -- \ \ \   & 86.9/\ \ -- \ \ \   & 82.0/\ \ -- \ \ \   & 67.2/\ \ -- \ \ \   \\
    GANomaly    \cite{akcay2018ganomaly} & 69.9/\ \ -- \ \ \   & 70.8/\ \ -- \ \ \   & 84.2/\ \ -- \ \ \   & 79.4/\ \ -- \ \ \   & 83.4/\ \ -- \ \ \   & 89.2/\ \ -- \ \ \   & 75.7/\ \ -- \ \ \   & 73.2/\ \ -- \ \ \   & 78.5/\ \ -- \ \ \   & 70.0/\ \ -- \ \ \   & 74.3/\ \ -- \ \ \   & 74.6/\ \ -- \ \ \   & 65.3/\ \ -- \ \ \   & 79.2/\ \ -- \ \ \   & 74.5/\ \ -- \ \ \   & 76.2/\ \ -- \ \ \   \\
    DSEBM  \cite{zhai2016deep} & 41.3/\ \ -- \ \ \   & 71.7/\ \ -- \ \ \   & 41.6/\ \ -- \ \ \   & 69.0/\ \ -- \ \ \   & 95.2/\ \ -- \ \ \   & 81.8/\ \ -- \ \ \   & 68.5/\ \ -- \ \ \   & 59.4/\ \ -- \ \ \   & 76.2/\ \ -- \ \ \   & 67.9/\ \ -- \ \ \   & 80.6/\ \ -- \ \ \   &  {99.9}/\ \ -- \ \ \   & 78.1/\ \ -- \ \ \   & 74.1/\ \ -- \ \ \   & 58.4/\ \ -- \ \ \   & 70.9/\ \ -- \ \ \   \\
%    1-NN   \cite{nazare2018pre} & 98.5  & 81.8  & 100.0  & 100.0  & 95.8  & 99.6  & 95.1  & 89.4  & 98.2  & 91.1  & 79.9  & 91.4  & 94.7  & 98.2  & 97.1  & 94.1  \\

    U-Student  \cite{bergmann2020uninformed} & 95.3/\ \ -- \ \ \   & 98.7/\ \ -- \ \ \   & 93.4/\ \ -- \ \ \   & 95.8/\ \ -- \ \ \   & 95.5/\ \ -- \ \ \   & 96.7/\ \ -- \ \ \   & 82.3/\ \ -- \ \ \   & 92.8/\ \ -- \ \ \   & 91.4/\ \ -- \ \ \   & 94.0/\ \ -- \ \ \   & 86.7/\ \ -- \ \ \   & 87.4/\ \ -- \ \ \   & 98.6/\ \ -- \ \ \   & 83.6/\ \ -- \ \ \   & 95.8/\ \ -- \ \ \   & 92.5/\ \ -- \ \ \   \\
    Patch SVDD  \cite{yi2020patch} & 92.9/92.6  & 94.6/96.2  & 90.9/97.4  & 97.8/91.4  & 96.5/90.8  & 98.6/98.1  & 90.3/96.8  & 76.7/95.8  & 92.0/97.5  & 90.9/98.0  & 86.1/95.1  & 81.3/95.7  & 100.0/98.1  & 91.5/97.0  & 97.9/95.1  & 92.1/95.7  \\
    DifferNet  \cite{rudolph2021same} & 92.9/\ \ -- \ \ \   & 84.0/\ \ -- \ \ \   & 97.1/\ \ -- \ \ \   & 99.4/\ \ -- \ \ \   & 99.8/\ \ -- \ \ \   & 99.0/\ \ -- \ \ \   & 95.9/\ \ -- \ \ \   & 86.9/\ \ -- \ \ \   & 99.3/\ \ -- \ \ \   & 96.1/\ \ -- \ \ \   & 88.8/\ \ -- \ \ \   & 96.3/\ \ -- \ \ \   & 98.6/\ \ -- \ \ \   & 91.9/\ \ -- \ \ \   & 95.1/\ \ -- \ \ \   & 94.9/\ \ -- \ \ \   \\
    VT-ADL  \cite{mishra2021vt} & 77.3/\ \ -- \ \ \   & 87.1/\ \ -- \ \ \   & 72.8/\ \ -- \ \ \   & 79.6/\ \ -- \ \ \   & 78.1/\ \ -- \ \ \   & 94.9/\ \ -- \ \ \   & 77.6/\ \ -- \ \ \   & 67.2/\ \ -- \ \ \   & 89.7/\ \ -- \ \ \   & 72.6/\ \ -- \ \ \   & 70.5/\ \ -- \ \ \   & 92.8/\ \ -- \ \ \   & 90.1/\ \ -- \ \ \   & 79.6/\ \ -- \ \ \   & 80.8/\ \ -- \ \ \   & 80.7/\ \ -- \ \ \   \\
    CutPaste  \cite{li2021cutpaste} & 93.9/98.3  &  {100.0}/97.5  &  {100.0}/ {99.5}  & 94.6/90.5  & 99.1/95.5  & 98.2/97.6  & 81.2/90.0  & 98.2/97.4  & 98.3/97.3  & 99.9/93.1  & 94.9/95.7  & 88.7/96.7  & 99.4/98.1  & 96.1/93.0  & 99.9/ {99.3}  & 96.1/96.0  \\
    PaDiM  \cite{defard2021padim} & \ \ -- \ \ /99.1  & \ \ -- \ \ /97.3  & \ \ -- \ \ /99.2  & \ \ -- \ \ /94.1  & \ \ -- \ \ /94.9  & \ \ -- \ \ /98.3  & \ \ -- \ \ /96.7  & \ \ -- \ \ /98.5  & \ \ -- \ \ /98.2  & \ \ -- \ \ /97.2  & \ \ -- \ \ /95.7  & \ \ -- \ \ /98.5  & \ \ -- \ \ /98.8  & \ \ -- \ \ /97.5  & \ \ -- \ \ /98.5  & 97.9/97.5  \\

    RFS energy  \cite{mansoor2021anomaly} & 98.4/\ \ -- \ \ \   & 89.6/\ \ -- \ \ \   &  {100.0}/\ \ -- \ \ \   & 96.9/\ \ -- \ \ \   & 98.1/\ \ -- \ \ \   &  {100.0}/\ \ -- \ \ \   & 92.0/\ \ -- \ \ \   & 89.4/\ \ -- \ \ \   & 99.9/\ \ -- \ \ \   & 98.2/\ \ -- \ \ \   & 94.5/\ \ -- \ \ \   & 70.0/\ \ -- \ \ \   & 99.2/\ \ -- \ \ \   & 91.9/\ \ -- \ \ \   & 98.7/\ \ -- \ \ \   & 94.5/\ \ -- \ \ \   \\
    Panda  \cite{reiss2021panda} &    \ \ -- \ \ /\ \ -- \ \ \    &    \ \ -- \ \ /\ \ -- \ \ \    &    \ \ -- \ \ /\ \ -- \ \ \    &   \ \ -- \ \ /\ \ -- \ \ \     &     \ \ -- \ \ /\ \ -- \ \ \   &    \ \ -- \ \ /\ \ -- \ \ \    &   \ \ -- \ \ /\ \ -- \ \ \     &    \ \ -- \ \ /\ \ -- \ \ \    &    \ \ -- \ \ /\ \ -- \ \ \    &     \ \ -- \ \ /\ \ -- \ \ \   &    \ \ -- \ \ /\ \ -- \ \ \    &   \ \ -- \ \ /\ \ -- \ \ \     &    \ \ -- \ \ /\ \ -- \ \ \    &    \ \ -- \ \ /\ \ -- \ \ \    &    \ \ -- \ \ /\ \ -- \ \ \    & 86.5/\ \ -- \ \ \   \\
    MKD  \cite{salehi2021multiresolution} &   79.3/95.6    &   78.0/91.8    &   95.0/98.0    &   91.6/82.8    &   94.3/84.8    &   99.4/96.3    &   89.2/82.4    &    80.5/95.9   &  98.4/94.6     &    73.5/86.4   &  82.7/89.6     &   83.3/96.0    &   92.2/96.1    &   85.5/76.5    &   93.2/93.9    & 87.7/90.7  \\

    DRAEM  \cite{zavrtanik2021draem} & 97.0/95.5  & 99.9/ {99.7}  &  {100.0}/98.6  & 99.6/ {99.2}  & 99.1/96.4  & 99.2/ {99.1}  & 91.8/94.7  &  {98.5}/94.3 &  {100.0}/ {99.7}  & 98.7/ {99.5}  &  {98.9}/97.6  & 93.9/97.6  &  {100.0}/98.1  & 93.1/90.9  &  {100.0}/98.8 & 98.0/97.3  \\
     FYD \cite{zheng2021focus} & 98.8/98.5  & 98.9/96.8  &  {100.0}/99.2  & 98.8/96.8  & 99.4/ {99.6}  &  {100.0}/98.3  & 95.3/ {97.5}  & 92.5/98.6 & 99.9/98.7  & 99.9/98.2 & 94.5/97.3 & 90.1/98.7 &  {100.0}/ {98.9}  & 99.2/ {98.1}  & 97.5/98.2 & 97.7/ {98.2}  \\
    CS-Flow  \cite{rudolph2021fully} &  {100.0}/\ \ -- \ \ \   & 99.0/\ \ -- \ \ \   &  {100.0}/\ \ -- \ \ \   &  {100.0}/\ \ -- \ \ \   &  {100.0}/\ \ -- \ \ \  & 99.8/\ \ -- \ \ \   &  {99.1}/\ \ -- \ \ \  & 97.1/\ \ -- \ \ \   & 99.6/\ \ -- \ \ \   & 99.1/\ \ -- \ \ \   & 98.6/\ \ -- \ \ \   & 97.6/\ \ -- \ \ \   & 91.9/\ \ -- \ \ \   & 99.3/\ \ -- \ \ \   & 99.7/\ \ -- \ \ \   & {98.7}/\ \ -- \ \ \ \\
    our & 98.9/ {98.9}  & 98.9/98.5  &   {100.0}/99.0  & 99.0/95.9  & 99.8/95.0  &   {100.0}/98.7  & 97.9/97.4  & 97.3/ {98.9} &  {100.0}/98.8 &   {100.0}/97.6 & 96.4/ {97.7}  & 99.0/ {99.4}  &  {100.0}/ {98.9}  &   {100.0}/94.7  & 99.3/98.8  & {99.1}/{97.9}  \\
   \hline
    \end{tabular}
   }

  \caption{Quantitative comparison with different methods on the MVTec dataset for anomaly detection/anomaly localization.
}

  \label{tab:01}
\end{table*}
\begin{algorithm}[h]
\caption{Discriminative Feature Learning Framework}
\label{alg:algorithm}
\textbf{Input}: train data $x_{train}$, test data $x_{test}$\\
\textbf{Output}: image anomaly score, pixel anomaly score\\
\textbf{Initialize} : hyper-parameters $\eta$, $c$, $\gamma$. pre-trained network $\phi$.
\begin{algorithmic}[1]
    \State   \textbf{Training}
\State $f_{i,l}= \phi (x_{train})$
\State Combine multi-level features using Eq.(2) to get $F_{i}$
\State Calculate the score map $S_{i}$ with eq.(3)
     \State $R \sim $uniform(0,1)
\If {$S_{i}^{j,k}>R_{i}^{j,k}$}
\State $\mathcal{M} = \mathcal{M}\bigcup F_{i}^{j,k}$
\EndIf
\While{$Loss\leq\eta*Loss_{start}$}
\State Training discriminative feature learning networks $\varphi$
\EndWhile
    \State $\mathcal{M}_{f} = \mathcal{M}_{f}\bigcup \varphi(v) ; v \in \mathcal{M}$
    \State \textbf{Testing}
 \State $f_{i,l}^{test} = \phi (x_{test})$
\State Combine multi-level features using Eq.(2) to get $F_{i}^{test}$
    \State $\mathcal{M}_{test} = \mathcal{M}_{test}\bigcup \varphi(v^{test}) ; v^{test} \in F_{i}^{test}$
    \State Calculate anomaly score using Eqs (10) and (11)
\State \textbf{return} image anomaly score, pixel anomaly score
\end{algorithmic}
\end{algorithm}
\section{Experiments}

\subsection{Dataset}
To evaluate the performance of our method,  we conduct extensive experiments on industrial and medical anomaly detection datasets.
Following, we describe these datasets in detail.

\textbf{MVTec AD dataset} \cite{bergmann2019mvtec} is one of the most challenging industrial anomaly detection datasets with multi-object and multi-defect-data.
It consists of 10 object and 5 texture categories, and each categorie contains 60-320 traing samples and about 100 test samples.

\textbf{Magnetic Tile Defects dataset} \cite{huang2020surface} (MTD) has a total of 1344 images, including 5 types of defects that are blowhole, crack, fray, break and uneven.
We use the same strategy as \cite{rudolph2021same} to divide the datasets into training set and test set.

\textbf{BeanTech AD dataset} \cite{mishra2021vt} (BTAD) contains 2540 images of three industrial products showcasing body and surface defects. All the images have been divided into training sets and test sets.

\textbf{Head CT dataset} \cite{salehi2021multiresolution} consists of 100 normal head CT images and 100 hemorrhagic head CT images.
We performe 5-folds cross-validation in this dataset, and use 80 normal images as training set and the rest as test set in each validation.

\subsection{Experimental settings}

We use ResNeXt-101 pretrained on ImageNet as our backbone network, and define the outputs of conv $2-5$ blocks as $1-4$ level features, respectively. In the following experinments, we use $1-3$ level features to calculate $F_{i}$.
Our MLP network $\varphi$ consists of two fully connected layers with output channel = 1024 and 512, respectively, and we use the Adam optimizer with a learning rate of $1 \times 10^{-4}$ to train it.
In addition, the hyperparameter $\eta$ of our early stop strategy is set to 0.8.
Furthermore, all images of the above datasets are resized to 224 \texttimes 224 pixels.
For anomaly detection and localization, we adopt area under the receiver operating characteristic curve (AU-ROC) as the  evaluation metric.

\subsection{Experimental results}
\subsubsection{Results on MVTec AD dataset} We compare our method with the state-of-the-art (SOTA) methods on  MVTec AD dataset, and the experimental results are shown in Table \ref{tab:01}. The results of GeoTrans, GANomaly and DSEBM are adopted from \cite{mansoor2021anomaly}, and the others are adopted from their original papers directly.
For anomaly detection, our method achieves the best performance in six categories, and surpasses the SOTA methods by 0.4\% in terms of the average AUROC (99.1\%). For anomaly localization, our method achieves SOTA performance in five categories and competitive performance in the average AUROC (97.9\%), which is only 0.3\% lower than FYD. Nevertheless, our method is 1.4\% higher than FYD in anomaly detection. In addition, we visualize the performance of our method for anomaly localization in Figure \ref{img3}.

\subsubsection{Results on MTD and BTAD datasets}
We evaluate our method on MTD and BTAD datasets. In Table \ref{tab:02}, we compare our method with GeoTrans, GANomaly, DSEBM and DifferNet on MTD dataset, and all experimental results are adopted from \cite{rudolph2021same}. Similarly, our method achieves best performance in the average AUROC (98.3\%), which is 0.6\% higher than the SOTA method.
For BTAD dataset, Table \ref{tab:03} presents quantitative comparison  with Panda, PaDiM and VT-ADL (results are adopted from its original paper). Our method achieves the average AUROC 93.3\%/97.3\% for anomaly detection/localization, which is 0.6\%/0.3\% improvement than the SOTA method.

\subsubsection{Results on Head CT dataset}
We conduct experiments on the Head CT dataset and the results are shown in Table \ref{tab:04}.
Our method achieves SOTA performance in the average AUROC (83.77\%), which outplays all SOTA methods by a huge margin (7.25\%).
All the experimental results validate that our proposed method is effective on anomaly detection and localization.

\begin{table}[htbp]
  \centering
  \resizebox {\linewidth}{!}{

    \begin{tabular}{llllll}
    \toprule
    \multicolumn{2}{c}{\multirow{2}[2]{*}{Category}} & our   & our   & our   & our \\
    \multicolumn{2}{c}{} & resnext101 & w\_resnet50 & resnet18 & efficientnet\_b4 \\
    \midrule
    \multicolumn{1}{c}{\multirow{5}[2]{*}{\begin{sideways}Textures\end{sideways}}} & Carpet & 98.9 / \textbf{98.9} & 98.6 / 98.3 & 97.4 / 98.3 & \textbf{99.5} / 98.4 \\
          & Grid  & 98.9 / \textbf{98.5} & 96.5 / 97.5 & 92.6 / 97.7 & \textbf{99.1} / 96.9 \\
          & Leather & \textbf{100 / 99.0} & 100 / 98.5 & 99.9 / 98.9 & 100 / 98.8 \\
          & Tile  & 99.0 / \textbf{95.9} & 98.9 / 93.9 & 98.5 / 93.3 & \textbf{99.4} / 93.4 \\
          & Wood  & \textbf{99.8 / 95.0} & 98.9 / 92.1 & 99.3 / 93.3 & 98.5 / 93.4 \\
    \midrule
    \multicolumn{1}{c}{\multirow{11}[2]{*}{\begin{sideways}Objects\end{sideways}}} & Bottle & \textbf{100 / 98.7} & 100 / 97.9 & 100 / 97.8 & 100 / 98.1 \\
          & Cable & 97.9 / 97.4 & 96.3 / 97.2 & 91.3 / 97.4 & \textbf{98.9 / 97.7} \\
          & Capsule & 97.3 / \textbf{98.9} & \textbf{98.1} / 98.6 & 95.9 / 98.6 & 95.0 / 98.7 \\
          & Hazelnut & \textbf{100 / 98.8} & 100 / 98.0 & 100 / 98.3 & 99.9 / 98.1 \\
          & MetalNut & \textbf{100 / 97.6} & 99.2 / 95.9 & 99.4 / 97.0 & 99.5 / 97.3 \\
          & Pill  & \textbf{96.2} / 97.7 & 93.7 / \textbf{98.7} & 95.3 / 98.2 & 93.2 / 97.2 \\
          & Screw & \textbf{99.0 / 99.4} & 97.2 / 98.2 & 98.3 / 99.3 & 97.2 / 99.0 \\
          & Toothbrush & \textbf{100 / 98.9} & 98.9 / 98.1 & 100 / 98.6 & 99.4 / 98.3 \\
          & Transistor & 100 / 94.7 & \textbf{100 / 97.4} & 98.9 / 96.1 & 98.4 / 97.8 \\
          & Zipper & \textbf{99.3 / 98.8} & 99.1 / 97.3 & 97.0 / 98.6 & 98.5 / 97.1 \\
          & Average & \textbf{99.1 / 97.9} & 98.4 / 97.2 & 97.6 / 97.4 & 98.4 / 97.3 \\
    \bottomrule
    \end{tabular}%
  }
    \caption{Anomaly detection and localization performance of different backbone networks on MVTec AD dataset in the format of (image level AUROC, pixel level AUROC).}
  \label{tab:05}%
\end{table}%
\begin{table} [htbp]
  \centering
 \scalebox {0.8}{
    \begin{tabular}{lrrrrr}
    \toprule
    Method & GeoTrans & GANomaly & DSEBM & DifferNet  & \textbf{ours} \\
    \midrule
    AUROC & 75.5  & 76.6  & 57.2  & 97.7   & \textbf{98.3} \\
    \bottomrule
    \end{tabular}%
 }
   \caption{Quantitative comparison with different methods on the MTD dataset for anomaly detection. }

  \label{tab:02}%
\end{table}%

\begin{table}[htbp]
  \centering
\scalebox {0.9}{
    \begin{tabular}{rrrrrr}

 \toprule
    Prdt & \tabincell{c}{Panda}&  \tabincell{c}{PaDiM} & VT-ADL & \tabincell{c} {OURS} \\
 \midrule
    1 & 96.4/96.4    & 99.4/97.2 &  \ \ -- \ \ / 99  & \textbf{99.7/97.4}  \\
   2 & 81.0/94.1    & 79.5/95.2    & \ \ -- \ \ / 94  &\textbf{86.1/95.8}  \\
   3 & 69.8/98.0    & \textbf{99.4/98.7}    & \ \ -- \ \ / 77 &94.2/98.6 \\
    Mean  & 82.4/96.2    &92.7/97.0& \ \ -- \ \ / 79    & \textbf{93.3/97.3}    \\
    \bottomrule
    \end{tabular}
}
  \caption{Quantitative comparison with different methods on the BTAD dataset for anomaly detection/localization. }

  \label{tab:03}
\end{table}
\begin{table}[htbp]
  \centering
 \scalebox {0.9}{
    \begin{tabular}{lrrrrrr}
    \toprule
    Method & GANomaly  & MKD & PaDiM & Panda & \textbf{ours} \\
    \midrule
    AUROC & 61.67  & 76.52 & 65.54 & 69.96 & \textbf{83.77} \\
    \bottomrule
    \end{tabular}%
 }
   \caption{Quantitative comparison with different methods on the Head CT dataset for anomaly detection.}

  \label{tab:04}%

\end{table}%

 \begin{figure}[htbp]
  \centering
  \includegraphics[width=0.5\textwidth]{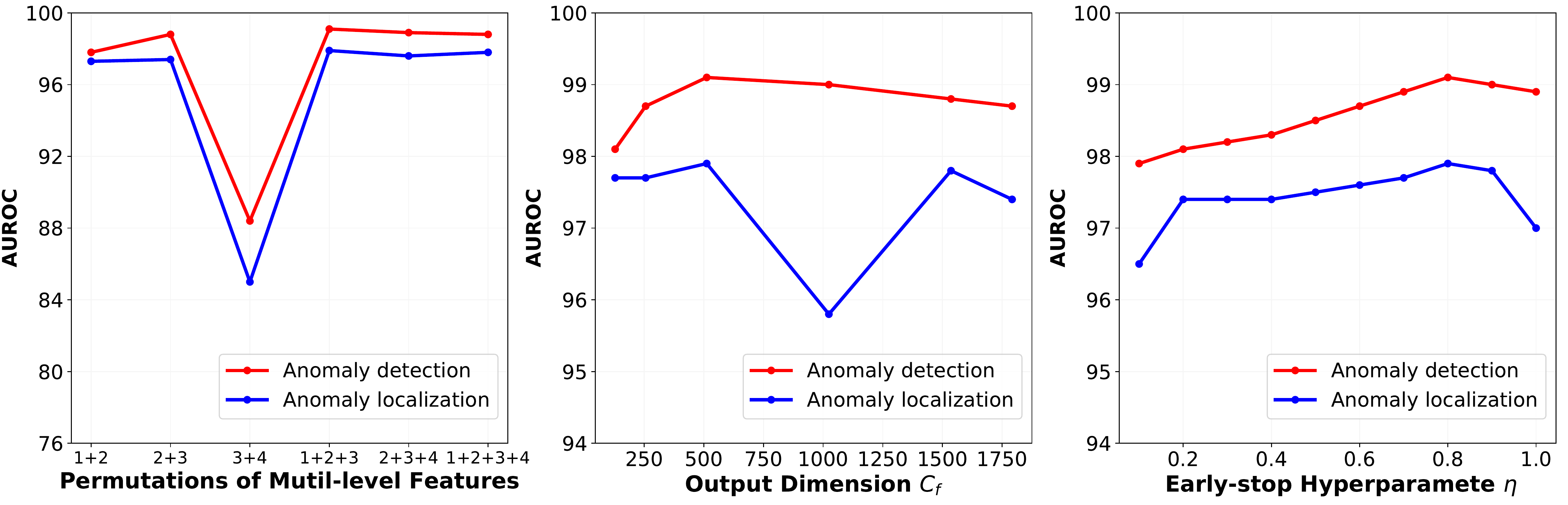} %1.png是图片文件的相对路径
  \caption{The experimental results of parameters analysis on MVTec AD dataset.} %caption是图片的标题
  \label{img4} %此处的label相当于一个图片的专属标志，目的是方便上下文的引用
\end{figure}

\begin{figure} [htbp]
  \centering
  \includegraphics[width=0.5\textwidth]{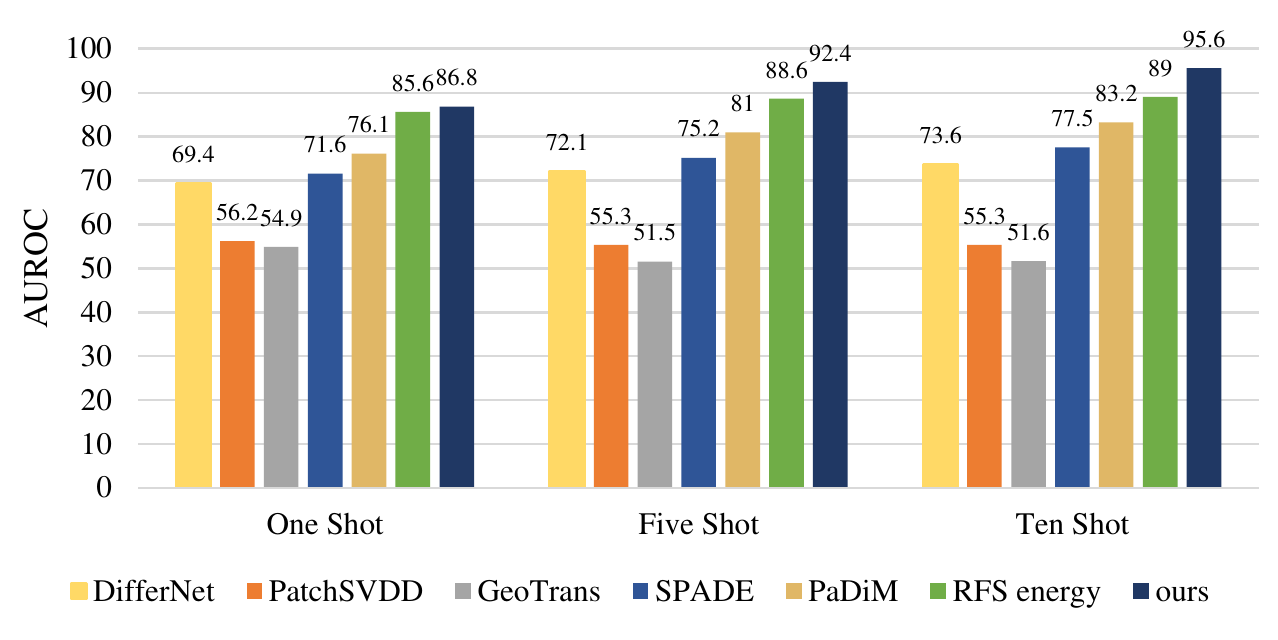} %1.png是图片文件的相对路径
  \caption{The average AUROC results on MVTec AD dataset for few shot anomaly detection.} %caption是图片的标题
  \label{img5} %此处的label相当于一个图片的专属标志，目的是方便上下文的引用
\end{figure}
\begin{table}[htbp]
  \centering
 \scalebox {0.9}{
    \begin{tabular}{lllrrr}
    \toprule
     P &  G & E & Anomaly detection & Anomaly localization \\
    \midrule
          & \checkmark     & \checkmark      &    79.3   & 87.7 \\
    \checkmark     &       & \checkmark      & 98.9  & 97.0 \\
    \checkmark     & \checkmark     &          & 98.1  & 97.8 \\
    \checkmark     & \checkmark    & \checkmark       & 99.1  & 97.9 \\
    \bottomrule
    \end{tabular}%
 }
  \caption {Quantitative results of our method for ablation studies  on MVTec AD dataset. P, G, E represent respectively feature extraction with
pre-trained network, gradient preference based feature selection, discriminative feature learning network using center
constraint}

  \label{tab:05}%

\end{table}%

\subsection{Discussion}

\subsubsection{Parameters analysis}

Four new parameters are introduced into our work, and we discuss their influence in this section. In the first place, we conduct comparative studies on our method with ResNeXt-101, Wide\_ResNet50, ResNet18, Efficientnet\_b4 as the pre-trained network $\phi$, and the results is shown in Table \ref{tab:06}.
Figure \ref{img4} presents the discussion of parameters on MVTec AD dataset.
The experimental results suggest that our mtheod achieves best performance when $\phi=$ ResNeXt-101, the permutation of mutil-level features $=1+2+3$, $C_{f}=512$, $\eta=0.8$.

\subsubsection{Ablation study}
We evaluate the contribution of each component of our proposed method by conducting a thorough ablation study, and present the quantitative results in Table \ref{tab:05}.
It turns out that the pretrained network has the greatest impact on the performance of our method because of high-quality features it can extract.
Additionally, gradient preference-based feature selection module is beneficial for the anomaly detection and localization on account of more representative feature vectors.
By combing three modules we designed, our proposed framework achieves encouraging results for both detection and localization.

\subsubsection{Few shot anomaly detection}
We further explore the effectiveness of our framework on few shot anomaly detection.
Figure \ref{img5} reports the compared results of our approach with DifferNet, PatchSVDD, GeoTrans, SPADE, PaDiM, and RFS.

\section{Conclusions}

In this paper, we proposed a novel discriminative feature learning framework with gradient preference for anomaly detection.
By constructing gradient preference-based feature selection strategy and discriminative feature learning network with center constraint, the features learned by pre-trained network can be brought close to better anomaly detection results.
Extensive experiments on industrial and medical anomaly detection datasets demonstrate that our proposed method achieves competitive results in both anomaly detection and localization.
In addition, our approach shows excellent superiority in few shots anomaly detection.
%\section{Acknowledgements}


\bibliographystyle{named}
\bibliography{ijcai22}

\begin{thebibliography}{}

\bibitem[\protect\citeauthoryear{Akcay \bgroup \em et al.\egroup
  }{2018}]{akcay2018ganomaly}
Samet Akcay, Amir Atapour-Abarghouei, and Toby~P Breckon.
\newblock Ganomaly: Semi-supervised anomaly detection via adversarial training.
\newblock In {\em ACCV}, pages 622--637, 2018.

\bibitem[\protect\citeauthoryear{Ak{\c{c}}ay \bgroup \em et al.\egroup
  }{2019}]{akccay2019skip}
Samet Ak{\c{c}}ay, Amir Atapour-Abarghouei, and Toby~P Breckon.
\newblock Skip-ganomaly: Skip connected and adversarially trained
  encoder-decoder anomaly detection.
\newblock In {\em IJCNN}, pages 1--8, 2019.

\bibitem[\protect\citeauthoryear{Ammar \bgroup \em et al.\egroup
  }{2021}]{mansoor2021anomaly}
Ammar, Amirali Khodadadian~Gostar, Alireza Bab-Hadiashar, and Reza
  Mansoor~Kamoona Hoseinnezhad.
\newblock Anomaly detection of defect using energy of point pattern features
  within random finite set framework.
\newblock {\em IEEE Transactions on Industrial Informatics}, 2021.

\bibitem[\protect\citeauthoryear{Bergmann \bgroup \em et al.\egroup
  }{2018}]{bergmann2018improving}
Paul Bergmann, Sindy L{\"o}we, Michael Fauser, David Sattlegger, and Carsten
  Steger.
\newblock Improving unsupervised defect segmentation by applying structural
  similarity to autoencoders.
\newblock {\em arXiv preprint arXiv:1807.02011}, 2018.

\bibitem[\protect\citeauthoryear{Bergmann \bgroup \em et al.\egroup
  }{2019}]{bergmann2019mvtec}
Paul Bergmann, Michael Fauser, David Sattlegger, and Carsten Steger.
\newblock Mvtec ad--a comprehensive real-world dataset for unsupervised anomaly
  detection.
\newblock In {\em CVPR}, pages 9592--9600, 2019.

\bibitem[\protect\citeauthoryear{Bergmann \bgroup \em et al.\egroup
  }{2020}]{bergmann2020uninformed}
Paul Bergmann, Michael Fauser, David Sattlegger, and Carsten Steger.
\newblock Uninformed students: Student-teacher anomaly detection with
  discriminative latent embeddings.
\newblock In {\em CVPR}, pages 4183--4192, 2020.

\bibitem[\protect\citeauthoryear{Defard \bgroup \em et al.\egroup
  }{2021}]{defard2021padim}
Thomas Defard, Aleksandr Setkov, Angelique Loesch, and Romaric Audigier.
\newblock Padim: a patch distribution modeling framework for anomaly detection
  and localization.
\newblock In {\em ICPR}, pages 475--489, 2021.

\bibitem[\protect\citeauthoryear{Golan and El-Yaniv}{2018}]{golan2018:deep}
Izhak Golan and Ran El-Yaniv.
\newblock Deep anomaly detection using geometric transformations.
\newblock {\em In Advances in Neural Information Processing Systems}, pages
  9758--9769, 2018.

\bibitem[\protect\citeauthoryear{Huang \bgroup \em et al.\egroup
  }{2020}]{huang2020surface}
Yibin Huang, Congying Qiu, and Kui Yuan.
\newblock Surface defect saliency of magnetic tile.
\newblock {\em The Visual Computer}, 36(1):85--96, 2020.

\bibitem[\protect\citeauthoryear{Li \bgroup \em et al.\egroup
  }{2018}]{li2018thoracic}
Zhe Li, Chong Wang, Mei Han, Yuan Xue, Wei Wei, Li-Jia Li, and Li~Fei-Fei.
\newblock Thoracic disease identification and localization with limited
  supervision.
\newblock In {\em CVPR}, pages 8290--8299, 2018.

\bibitem[\protect\citeauthoryear{Li \bgroup \em et al.\egroup
  }{2020}]{li2020multi}
Yingying Li, Jie Wu, Xue Bai, Xipeng Yang, Xiao Tan, Guanbin Li, Shilei Wen,
  Hongwu Zhang, and Errui Ding.
\newblock Multi-granularity tracking with modularlized components for
  unsupervised vehicles anomaly detection.
\newblock In {\em CVPRW}, pages 586--587, 2020.

\bibitem[\protect\citeauthoryear{Li \bgroup \em et al.\egroup
  }{2021}]{li2021cutpaste}
Chun-Liang Li, Kihyuk Sohn, Jinsung Yoon, and Tomas Pfister.
\newblock Cutpaste: Self-supervised learning for anomaly detection and
  localization.
\newblock In {\em CVPR}, pages 9664--9674, 2021.

\bibitem[\protect\citeauthoryear{Liu \bgroup \em et al.\egroup
  }{2020}]{liu2020towards}
Wenqian Liu, Runze Li, Meng Zheng, Srikrishna Karanam, Ziyan Wu, Bir Bhanu,
  Richard~J Radke, and Octavia Camps.
\newblock Towards visually explaining variational autoencoders.
\newblock In {\em CVPR}, pages 8642--8651, 2020.

\bibitem[\protect\citeauthoryear{Mishra \bgroup \em et al.\egroup
  }{2021}]{mishra2021vt}
Pankaj Mishra, Riccardo Verk, Daniele Fornasier, Claudio Piciarelli, and
  Gian~Luca Foresti.
\newblock Vt-adl: A vision transformer network for image anomaly detection and
  localization.
\newblock {\em arXiv preprint arXiv:2104.10036}, 2021.

\bibitem[\protect\citeauthoryear{Nguyen \bgroup \em et al.\egroup
  }{2019}]{nguyen2019anomaly}
Duc~Tam Nguyen, Zhongyu Lou, Michael Klar, and Thomas Brox.
\newblock Anomaly detection with multiple-hypotheses predictions.
\newblock In {\em ICML}, pages 4800--4809, 2019.

\bibitem[\protect\citeauthoryear{Reiss \bgroup \em et al.\egroup
  }{2021}]{reiss2021panda}
Tal Reiss, Niv Cohen, Liron Bergman, and Yedid Hoshen.
\newblock Panda: Adapting pretrained features for anomaly detection and
  segmentation.
\newblock In {\em CVPR}, pages 2806--2814, 2021.

\bibitem[\protect\citeauthoryear{Rudolph \bgroup \em et al.\egroup
  }{2021}]{rudolph2021same}
Marco Rudolph, Bastian Wandt, and Bodo Rosenhahn.
\newblock Same same but differnet: Semi-supervised defect detection with
  normalizing flows.
\newblock In {\em WACV}, pages 1907--1916, 2021.

\bibitem[\protect\citeauthoryear{Rudolph \bgroup \em et al.\egroup
  }{2022}]{rudolph2021fully}
Marco Rudolph, Tom Wehrbein, Bodo Rosenhahn, and Bastian Wandt.
\newblock Fully convolutional cross-scale-flows for image-based defect
  detection.
\newblock In {\em WACV}, 2022.

\bibitem[\protect\citeauthoryear{Salehi \bgroup \em et al.\egroup
  }{2021}]{salehi2021multiresolution}
Mohammadreza Salehi, Niousha Sadjadi, Soroosh Baselizadeh, Mohammad~H Rohban,
  and Hamid~R Rabiee.
\newblock Multiresolution knowledge distillation for anomaly detection.
\newblock In {\em CVPR}, pages 14902--14912, 2021.

\bibitem[\protect\citeauthoryear{Schlegl \bgroup \em et al.\egroup
  }{2019}]{schlegl2019f}
Thomas Schlegl, Philipp Seeb{\"o}ck, Sebastian~M Waldstein, Georg Langs, and
  Ursula Schmidt-Erfurth.
\newblock f-anogan: Fast unsupervised anomaly detection with generative
  adversarial networks.
\newblock {\em Medical image analysis}, 54:30--44, 2019.

\bibitem[\protect\citeauthoryear{Sheynin \bgroup \em et al.\egroup
  }{2021}]{sheynin2021hierarchical}
Shelly Sheynin, Sagie Benaim, and Lior Wolf.
\newblock A hierarchical transformation-discriminating generative model for few
  shot anomaly detection.
\newblock {\em arXiv preprint arXiv:2104.14535}, 2021.

\bibitem[\protect\citeauthoryear{Yi and Yoon}{2020}]{yi2020patch}
Jihun Yi and Sungroh Yoon.
\newblock Patch svdd: Patch-level svdd for anomaly detection and segmentation.
\newblock In {\em ACCV}, 2020.

\bibitem[\protect\citeauthoryear{Zavrtanik \bgroup \em et al.\egroup
  }{2021}]{zavrtanik2021draem}
Vitjan Zavrtanik, Matej Kristan, and Danijel Skocaj.
\newblock Draem-a discriminatively trained reconstruction embedding for surface
  anomaly detection.
\newblock In {\em ICCV}, pages 8330--8339, 2021.

\bibitem[\protect\citeauthoryear{Zhai \bgroup \em et al.\egroup
  }{2016}]{zhai2016deep}
Shuangfei Zhai, Yu~Cheng, Weining Lu, and Zhongfei Zhang.
\newblock Deep structured energy based models for anomaly detection.
\newblock In {\em ICML}, pages 1100--1109, 2016.

\bibitem[\protect\citeauthoryear{Zheng \bgroup \em et al.\egroup
  }{2021}]{zheng2021focus}
Ye~Zheng, Xiang Wang, Rui Deng, Tianpeng Bao, Rui Zhao, and Liwei Wu.
\newblock Focus your distribution: Coarse-to-fine non-contrastive learning for
  anomaly detection and localization.
\newblock {\em arXiv preprint arXiv:2110.04538}, 2021.

\end{thebibliography}

\end{document}